\documentclass[10pt,twocolumn,letterpaper]{article}

\usepackage{cvpr}
\usepackage{times}
\usepackage{epsfig}
\usepackage{graphicx}
\usepackage{amsmath}
\usepackage{amssymb}

\usepackage{color}
\usepackage{makecell}
\usepackage{multirow}
\usepackage{CJKutf8}

\usepackage[breaklinks=true,bookmarks=false]{hyperref}

\cvprfinalcopy 

\def\cvprPaperID{****} 

\begin{document}
\begin{CJK}{UTF8}{gbsn}

\title{Towards Accurate Scene Text Recognition with Semantic Reasoning Networks}

\author{Deli Yu$^{13}$\thanks{Equal contribution. This work is done when Deli Yu is an intern at Baidu Inc.}\quad Xuan Li$^{2*}$\quad Chengquan Zhang$^{2*}$\thanks{Corresponding author.}\quad Tao Liu$^2$\\
\quad Junyu Han$^2$\quad Jingtuo Liu$^2$\quad Errui Ding$^2$\\
School of Artificial Intelligence, University of Chinese Academy of Sciences$^1$ \\
Department of Computer Vision Technology(VIS), Baidu Inc.$^2$\\
National Laboratory of Pattern Recognition, Institute of Automation, Chinese Academy of Sciences$^3$\\
{\tt\small yudeli2018@ia.ac.cn}\\
{\tt\small $\{$lixuan12, zhangchengquan, liutao32, hanjunyu, liujingtuo, dingerrui$\}$@baidu.com}
}




\maketitle

\begin{abstract}
Scene text image contains two levels of contents: visual texture and semantic information.
Although the previous scene text recognition methods have made great progress over the past few years, the research on mining semantic information to assist text recognition attracts less attention, only RNN-like structures are explored to implicitly model semantic information.
However, we observe that RNN based methods have some obvious shortcomings, such as time-dependent decoding manner and one-way serial transmission of semantic context, which greatly limit the help of semantic information and the computation efficiency.
To mitigate these limitations, we propose a novel end-to-end trainable framework named semantic reasoning network (SRN) for accurate scene text recognition, where a global semantic reasoning module (GSRM) is introduced to capture global semantic context through multi-way parallel transmission. 
The state-of-the-art results on 7 public benchmarks, including regular text, irregular text and non-Latin long text, verify the effectiveness and robustness of the proposed method. 
In addition, the speed of SRN has significant advantages over the RNN based methods, demonstrating its value in practical use.
\end{abstract}
\vspace{-2mm}

\section{Introduction}\label{sec:intro}
\begin{figure}[!h]
\begin{center}
\includegraphics[width=0.9\linewidth]{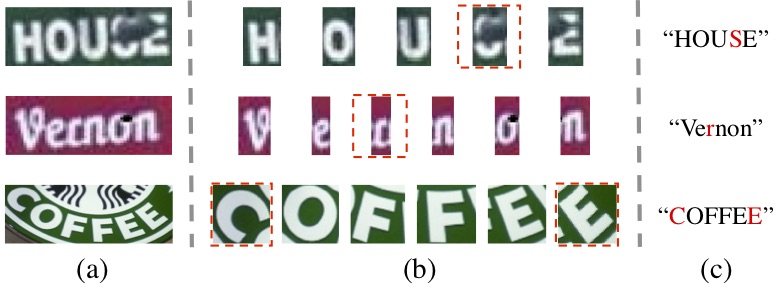}
\end{center}
\vspace{-3mm}
\caption{Examples of text in the wild. (a) are some difficult scene text images, (b) are individual characters extracted separately from (a), and (c) are the corresponding semantic word contents. The characters with red dashed boxes in (b) are easy to be misclassified, only based on visual features.}
\label{fig:intro_case}
\vspace{-5mm}
\end{figure}

Text has rich semantic information, which has been used in many computer vision based applications such as automatic driving~\cite{yu2019videotext}, travel translator~\cite{wu2019editing}, product retrieval, etc. 
Scene text recognition is a crucial step in scene text reading system.
Although sequence-to-sequence recognition has made several remarkable breakthroughs in the past decades~\cite{lee2016recursive,wojna2017attention-2D,yang2017learning-2D}, text recognition in the wild is still a big challenge, caused by the significant variations of scene text in color, font, spatial layout and even uncontrollable background.

Most of the recent works have attempted to improve the performance of scene text recognition from the perspective of extracting more robust and effective visual features, such as upgrading the backbone networks~\cite{cheng2017focusing,liao2019mask,shi2018aster}, adding rectification modules~\cite{shi2016robust,shi2018aster,yang2019symmetry,zhan2019esir} and improving attention mechanisms~\cite{cheng2017focusing,wojna2017attention-2D,yang2017learning-2D}.  
Nevertheless, it is a fact that, for a human, the recognition of scene text is not only dependent on visual perception information, but also affected by the high-level text semantic context understanding.
As some examples shown in Fig.~\ref{fig:intro_case}, it is very difficult to distinguish each character in those images separately when only visual features are considered, especially the characters highlighted with red dotted boxes. 
Instead, taking semantic context information into consideration, human is likely to infer the correct result with the total word content.
\begin{figure}[!h]
\begin{center}
\includegraphics[width=0.9\linewidth]{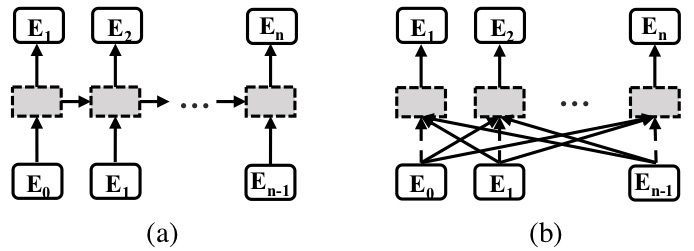}
\end{center}
\vspace{-5mm}
\caption{Two different manners of semantic context delivery. (a) is one-way serial transmission, and (b) is multi-way parallel transmission.}
\label{fig:intro_sem}
\vspace{-5mm}
\end{figure}

Unfortunately, for the semantic information, the mainstream text recognition methods consider it in the way of one-way serial transmission, such as 
\cite{cheng2017focusing,shi2016robust,shi2018aster,wojna2017attention-2D,yang2019symmetry,yang2017learning-2D,zhan2019esir}, which recursively perceive the character semantic information of the last decoding time step, as shown in Fig.~\ref{fig:intro_sem} (a). 
There are several obvious drawbacks in this manner: First, it can only perceive very limited semantic context from each decoding time step, even no useful semantic information for the first decoding time step. Second, it may pass the wrong semantic information down and cause a error accumulation when the wrong decoding is raised at an earlier time step.
Meanwhile, the serial mode is hard to be paralleled, thus it is always time-consuming and inefficient.

In this paper, we introduce a sub-network structure named global semantic reasoning module (GSRM) to tackle these disadvantages. The GSRM considers global semantic context in a novel manner of multi-way parallel transmission. As is shown in Fig.~\ref{fig:intro_sem} (b), the multi-way parallel transmission can simultaneously perceive the semantic information of all characters in a word or text line, which is much more robust and effective. Besides, the wrong semantic content of the individual character can only cause quite limited negative impact on other steps.

Furthermore, we propose a novel framework named semantic reasoning network (SRN) for accurate scene text recognition, which integrates not only global semantic reasoning module (GSRM) but also parallel visual attention module (PVAM) and visual-semantic fusion decoder (VSFD). 
The PVAM is designed to extract visual features of each time step in a parallel attention mechanism, and the VSFD aims to develop an effective decoder with the combination of visual information and semantic information. 
The effectiveness and robustness of the SRN are confirmed by extensive experiments, which are discussed in Sec.~\ref{sec:experiment}.

The major contributions of this paper are threefold.
First, we propose a global semantic reasoning module (GSRM) to consider global semantic context information, which is more robust and efficient than one-way serial semantic transmission methods.
Second, a novel framework named semantic reasoning network (SRN) for accurate scene text recognition is proposed, which combines both visual context information and semantic context information effectively.
Third, SRN can be trained in an end-to-end manner, and achieve the state-of-the-art performance on several benchmarks including regular text, irregular text and non-Latin long text.

\section{Related Work}
\begin{figure*}[htp]
  \begin{center}
  \includegraphics[width=0.95\textwidth]{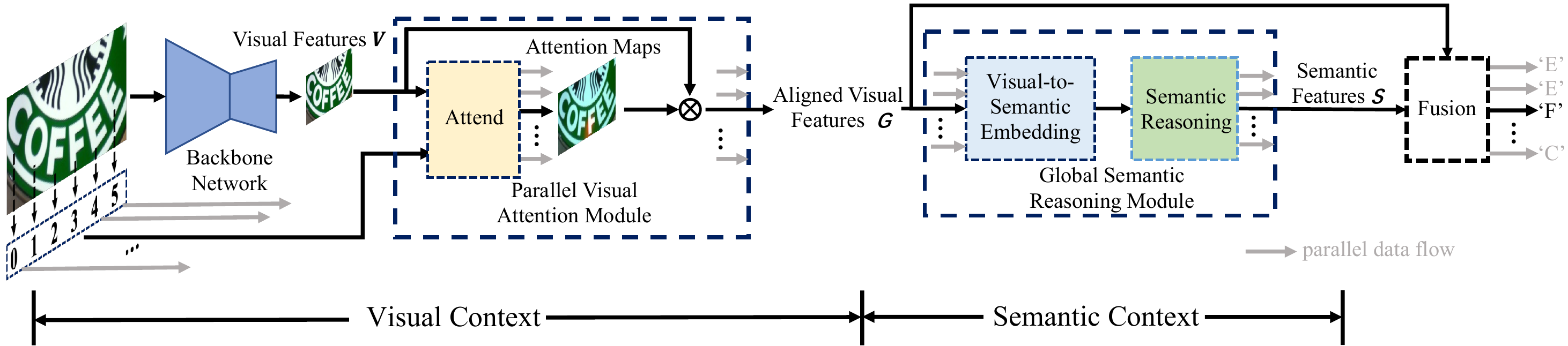}
  \vspace{-3mm}
  \end{center}
      \caption{The pipeline of the semantic reasoning network (SRN). 
      }
  \label{fig:pipeline}
  \vspace{-3mm}
  \end{figure*}
The research of scene text recognition has made significant progress in the past few years. 
In this section, we will first review some recent text recognition methods, and summarize them into two categories: semantic context-free and semantic context-aware, according to whether semantic information is used. Besides, some popular context modeling structures will also be presented.

\textbf{Semantic context-free methods} regard scene text recognition as a purely visual classification task, without using any semantic information explicitly. CRNN~\cite{crnn-shi2016end} firstly combined CNN and RNN to extract sequential visual features of a given text image, and then directly fed them into a CTC decoder to predict the best character category of each time step, where CTC~\cite{graves2006CTC} only maximized the probability of all the paths that can reach the ground truth according to the visual classification of each position.
In order to alleviate the back-propagating computation burden of CTC loss, Xie~\etal~\cite{xie2019aggregation} proposed the aggregation cross-entropy (ACE) loss to optimize the statistical frequency of each character along the time dimension, enhancing the efficiency greatly.
Inspired by the success of visual segmentation, Liao~\etal~\cite{liao2019scene-fcn} used FCN to predict the character categories of each position by pixel-level classification, and gathered characters into text lines by heuristic rules. However, this method requires expensive character-level annotation. Instead of optimizing the decoding accuracy of each step, Jaderberg~\cite{synth90k-1-Jaderberg14c} directly used CNNs to classify 90k kinds of text image, each of which represented a word.
In general, these methods ignore to take semantic context into account.

\textbf{Semantic context-aware methods} try to capture semantic information to assist the scene text recognition. Most of those methods follow the one-way semantic transmission manner, for example, Lee~\etal~\cite{lee2016recursive} encoded the input text image horizontally into 1D sequential visual features, and then guided visual features to attend the corresponding region with the help of semantic information of last time step.
As we mentioned before, some of the latest works focus on how to mine more effective visual features, especially for irregular text. 
In order to eliminate the negative effects brought by perspective distortion and distribution curvature, Shi \etal~\cite{shi2018aster} added a rectification module before sequence recognition, in which a spatial transform network~\cite{jaderberg2015spatial} with multiple even control point pairs was adopted. 
Zhan \etal~\cite{zhan2019esir} employed a line fitting transformation with iterative refinement mechanism to rectify the irregular text image.
Furthermore, Yang \etal~\cite{yang2019symmetry} proposed a symmetry-constrained rectification network based on the rich local attributes to generate better rectification results.
There are some methods alleviating the challenge of irregular text recognition by enhancing spatial visual features. 
Cheng~\etal~\cite{cheng2018aon} extracted scene text features in four directions and designed a filter gate to control the contribution of features from each direction. 
Wojna~\etal~\cite{wojna2017attention-2D} introduced additional encoding of spatial coordinates on 2D feature maps to increase the sensitivity to sequential order.
However, these works do not fully exploit semantic context information, which is exactly what we want to focus on in this paper.

\textbf{Context modeling structures} are designed to capture information in a certain time or spatial range.
RNN is good at capturing dependencies of sequence data, but its inherent sequential behavior hinders parallel computation~\cite{wojna2017attention-2D} during the training and inference.
To solve those issues, ByteNet~\cite{kalchbrenner2016neural} and ConvS2S~\cite{gehring2017convolutional} directly used CNNs as encoder. These methods can be fully parallelized during training and inference to make better use of the hardware, but cannot flexibly capture the global relations, due to the limitation of receptive field size.
Recently, the structure of transformer~\cite{lin2017structured-selfattention} has been proposed to capture global dependencies and relate two signals at arbitrary positions with constant computation complexity. 
In addition, transformer has been proved to be effective in many tasks of computer vision~\cite{hu2018relation,wang2018non} and natural language processing~\cite{vaswani2017attention}.
In this paper, we not only adopt transformer to enhance the visual encoding features, but also use the similar structure to reason semantic content.
\section{Approach}
The SRN is an end-to-end trainable framework that consists of four parts: backbone network, parallel visual attention module (PVAM), global semantic reasoning module (GSRM), and visual-semantic fusion decoder (VSFD).
Given an input image, the backbone network is first used to extract 2D features $V$. Then, the PVAM is used to generate $N$ aligned 1-D features $G$, where each feature corresponds to a character in the text and captures the aligned visual information. These $N$ 1-D features $G$ are then fed into our GSRM to capture the semantic information $S$. Finally, the aligned visual features $G$ and the semantic information $S$ are fused by the VSFD to predict $N$ characters. For text string shorter than $N$, 'EOS' are padded. The detailed structure of SRN is shown in Fig.~\ref{fig:pipeline}.

\subsection{Backbone Network}\label{sec:backbone network}
We use FPN~\cite{lin2017feature} to aggregate hierarchical feature maps from the stage-3, stage-4 and stage-5 of ResNet50~\cite{he2016deep} as the backbone network. Thus, the feature map size of ResNet50+FPN is $1/8$ of the input image, and the channel number is $512$.
Inspired by the idea of non-local mechanisms~\cite{buades2005non}, we also adopt the transformer unit~\cite{vaswani2017attention} which is composed of a positional encoding, multi-head attention networks and a feed-forward module to effectively capture the global spatial dependencies. 
2D feature maps are fed into two stack transformer units, where the number of heads in multi-head attention is 8 and the feed-forward output dimension is $512$.
After that, the final enhanced 2D visual features are extracted, denoted as $V$, $v_{ij} \in \mathbf{R}^{d}$, where $d=512$.

\subsection{Parallel Visual Attention Module}
Attention mechanism is widely used in sequence recognition~\cite{cheng2017focusing,shi2016robust}. It can be regarded as a form of feature alignment where relevant information in the input is aligned to the corresponding output. 
Therefore, attention mechanism is used to generate $N$ features where each feature corresponds to a character in the text.
Existing attention based methods are inefficient because of some time-dependent terms. In this work, a new attention method named parallel visual attention (PVA) is introduced to improve the efficiency by breaking down these barriers.

Generally, attention mechanism can be described as follows: Given a key-value set ${(k_i, v_i)}$ and a query $q$,  the similarities between the query $q$ and all keys ${k_i}$ are computed. Then the values ${v_i}$ are aggregated according to the similarities. Specifically, in our work, the key-value set is the input 2D features ${(v_{ij}, v_{ij})}$.  
Following the Bahdanau attention~\cite{bahdanau2014neural}, the existing methods use the hidden state $H_{t-1}$ as the query to generate the $t$-th feature. To make the computation parallel, the reading order is used as the query instead of the time-dependent term $H_{t-1}$.
The first character in the text has reading order $0$. The second character has reading order $1$, and etc. Our parallel attention mechanism can be summarized as:
\begin{equation} 
\left \{
\begin{array}{cl}
e_{t,ij} &= W_e^T tanh(W_o f_{o}(O_t)+ W_v v_{ij}) \\
\alpha_{t,{ij}} &= \frac{\displaystyle \exp(e_{t,{ij}})} {\displaystyle \sum_{\forall i,j}\exp(e_{t,{ij}})} 
\end{array}
\right .
\label{eq:attn}
\end{equation}
where, $W_e$, $W_o$, and $W_v$ are trainable weights. $O_t$ is the character reading order whose value is in the list of $[0,1,...,N-1]$, and $f_{o}$ is the embedding function.

Based on the idea of PVA, we design the parallel visual attention module (PVAM) to align all visual features of all time steps. 
The aligned visual feature of the $t$-th time step can be represented as:
\begin{equation}
    g_t = \sum_{\forall i,j} \alpha_{t,ij} v_{ij}
\end{equation}
Because the calculation method is time independent, PVAM outputs aligned visual features ($G$, $g_{t} \in \mathbf{R}^{d}$) of all time steps in parallel.

\begin{figure}[!h]
\begin{center}
\includegraphics[width=0.98\linewidth]{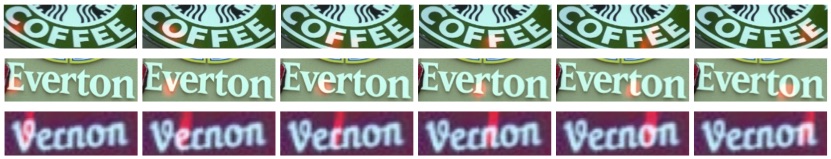}
\end{center}
\vspace{-3mm}
\caption{Attention maps calculated by PVAM.}
\label{fig:attenmap}
\vspace{-5mm}
\end{figure}

As some attention maps shown in Fig.~\ref{fig:attenmap}, the obtained attention maps can pay attention to the visual areas of corresponding characters correctly and the effectiveness of PVAM is verified well. 

\subsection{Global Semantic Reasoning Module}
In this section, we propose the global semantic reasoning module (GSRM) that follows the idea of multi-way parallel transmission to overcome the drawbacks of one-way semantic context delivery.
Firstly, we review the probability formula to be maximized in the Bahdanau attention mechanism, a typical RNN-like structure. It can be expressed as the following:
\begin{equation}
\begin{aligned}
    p(y_{1}y_{2}\cdots y_{N}) = \prod\limits_{t=1}^{N}p(y_{t}|e_{t-1},H_{t-1},g_t)
\end{aligned}
\label{formula:att}
\end{equation}
where $e_t$ is regarded as the word embedding of $t$-th label $y_t$. At each time step, RNN-like methods can refer to the previous labels or predicted results during the training or inference. So they work in a sequential way since the previous information like $e_{t-1}$ and $H_{t-1}$ can only be captured at time step $t$, which  limits the ability of semantic reasoning
and causes low efficiency during inference. 

To overcome the aforementioned problems, our key insight is that instead of using the real word embedding $e$, we use an approximated embedding $e'$ which is time-independent.
Several benefits can be made from this improvement. 
1) First, the hidden states value of last steps $H_{t-1}$ are able to be removed from the Eq.~\ref{formula:att} and thus the serial forward process will be upgraded to a parallel one with high efficiency because all time-dependent terms are eliminated. 
2) Second, the global semantic information, including all the former and the latter characters, is allowed to be combined together and to reason the appropriate semantic content of the current time.
Hence, the probability expression can be upgraded as:
\begin{equation}
\begin{aligned}
\begin{array}{cl}
p(y_{1}y_{2}\cdots y_{N})= \prod\limits_{t=1}^{N}p(y_{t}|f_r(e_{1}\cdots e_{t-1}e_{t+1}\cdots e_{N}), g_t) \\ 
\approx \prod\limits_{t=1}^{N}p(y_{t}|f_r(e'_{1}\cdots e'_{t-1}e'_{t+1}\cdots e'_{N}), g_t)
\end{array}
\label{formula:approx}
\end{aligned}
\end{equation}
where $e'_t$ is the approximate embedding information of $e_t$ at the $t$-th time step. The $f_r$ in Eq.~\ref{formula:approx} means a function that can build the connection between the global semantic context and current semantic information. If we denote the $s_{t} = f_r(e_{1}\cdots e_{t-1}e_{t+1}\cdots e_{N})$ and $s_{t}$ is the $t$-th feature of semantic information $S$, the Eq.~\ref{formula:approx} can be simplified to the following one:
\begin{equation}
\begin{aligned}
\begin{array}{cl}
p(y_{1}y_{2}\cdots y_{N})\approx \prod\limits_{t=1}^{N}p(y_{t}|s_{t}, g_t)
\end{array}
\label{formula:simplified}
\end{aligned}
\end{equation}
Inheriting from the above spirit, we propose the GSRM, by which the function $f_r$ in Eq.~\ref{formula:approx} is modeled, to make the supposition come true and benefit from it.
The structure of GSRM is composed of two key parts: visual-to-semantic embedding block and semantic reasoning block.

\begin{figure}
    \centering
    \includegraphics[width=0.98\columnwidth]{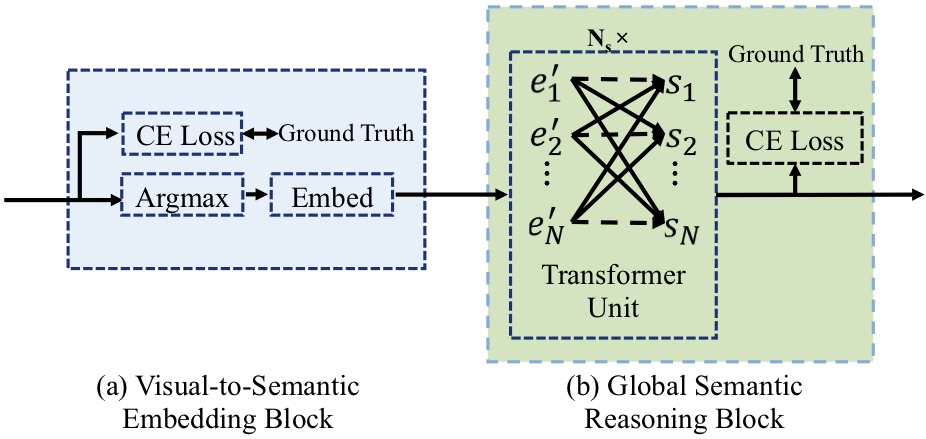}
    \caption{The detailed structure of GSRM.}
    \label{fig:GSRM}
    \vspace{-3mm}
\end{figure}

\textbf{Visual-to-semantic embedding block} is used for the purpose of generating $e'$, and the detailed structure is shown in Fig.~\ref{fig:GSRM} (a). Thanks to the PVAM, the features we get are already aligned to every time step or every target character. The aligned visual features $G$ are fed to a fully-connection layer with softmax activation first, and the embedding loss $L_e$, where cross entropy loss is utilized, is added to make them more concentrate on the target characters.
\begin{equation}
L_e = - \frac{1}{N}\sum\limits_{t=1}^{N} \log p(y_{t}|g_{t})
\label{eq:prob_pesudo}
\end{equation}
Next, embedding vector $e'_t$ is calculated based on the most likely output characters of $g_t$ by the $argmax$ operation and an embedding layer. 

\textbf{Semantic reasoning block} serves to realize the global semantic reasoning, that is to model the function $f_r$ in Eq.~\ref{formula:approx}. Specially, the structure of GSRM is illustrated in Fig.~\ref{fig:GSRM} (b). 
Several transformer units are followed with the masked $e'$ allowing the model to perceive the global context information with high efficiency. Meanwhile, both first-order relations and higher-order relations, such as word semantic, can be implicitly modeled by multiple transformer units.
Finally, the semantic features of every step is output through this module, which is defined as $S$, $s_t\in \mathbf{R}^{d}$, $d=512$. Meanwhile, the reasoning loss $L_r$ is added on the $s$. The objective function can be defined as
\begin{equation}
L_r = - \frac{1}{N}\sum\limits_{t=1}^{N} \log p(y_{t}|s_{t})
\label{eq:prob_ngram}
\end{equation}
The cross entropy loss is performed to optimize the objective probability from the perspective of semantic information, which also helps to reduce the convergence time. 
It is noticeable that the global semantic is reasoned in a parallel way in the GSRM, making SRN run much faster than the traditional attention based methods, especially in the cases of long text.

\subsection{Visual-Semantic Fusion Decoder} \label{sec:Visual-Semantic Fusion Decoder}
As we discussed in the Sec.~\ref{sec:intro}, it is very important to consider both visual aligned features $G$ and semantic information $S$ for scene text recognition.
However, $G$ and $S$ belong to different domains, and their weights for the final sequence recognition in different cases should be different.
Inspired by the gated unit~\cite{arevalo2017gated}, we introduce some trainable weights to balance the contribution of features from different domains in our VSFD.
The operation can be formulated as the following:
\begin{equation}
  \left\{
      \begin{array}{cl}
         z_t &=\sigma (\mathbf{W_z} \cdot [g_t,s_t])\\ 
         f_t &= z_t *g_t + (1-z_t)* s_t \\
      \end{array}
  \right .
  \label{eq:fusion}
\end{equation}
where $W_z$ is trainable weight, $f_t$ is the $t$-th fused feature vector, $t\in [1, N]$. All fused features can be denoted as $F$, $f_{t} \in \mathbf{R}^{d}$, and are used to predict the final characters in a non-recursive manner, the objective function is as follows:
\begin{equation}
L_f = - \frac{1}{N}\sum\limits_{t=1}^{N} \log p(y_{t}|f_{t})
\label{eq:prob_fusion}
\end{equation}

Combing all constraint functions in GSRM and VSFD, the final objective function is summarized as follows:
\begin{equation}
      Loss =   \alpha_e L_e+ \alpha_r L_r+ \alpha_f L_f
\end{equation}
where $L_e$, $L_r$ and $L_f$ represent embedding loss, reasoning loss and final decoder loss, respectively. The weights of $\alpha_e$, $\alpha_r$ and $\alpha_f$ are set to 1.0, 0.15 and 2.0 to trade off these three constraints. 

\section{Experiment}\label{sec:experiment}

\subsection{Datasets}
There are six Latin scene text benchmarks chosen to evaluate our method.

\textbf{ICDAR 2013}~(IC13)~\cite{icd2013-karatzas2013icdar} contains
1095 testing images. Using the protocol of \cite{svt-wang2011end}, we discard images that contain non-alphanumeric characters or less than three characters.

\textbf{ICDAR 2015}~(IC15)~\cite{icd2015-karatzas2015icdar} is taken with Google Glasses without careful position and focusing. 
We follow the same protocol of ~\cite{cheng2017focusing} and use only 1811 test images for evaluation without some extremely distorted images.

\textbf{IIIT 5K-Words}~(IIIT5k)~\cite{iiit5k-MishraBMVC12} is collected from the website and comprises of 3000 testing images.

\textbf{Street View Text}~(SVT)~\cite{svt-wang2011end} has
647 testing images cropped form Google Street View. Many images are severely corrupted by noise, blur, and low resolution. 

\textbf{Street View Text-Perspective}~(SVTP)~\cite{svtp-quy2013recognizing} is also cropped form Google Street View. There are 639 test images in this set and many of them are perspectively distorted. 

\textbf{CUTE80}~(CUTE) is proposed in \cite{cute-risnumawan2014robust} for curved text recognition.
288 testing images are cropped from full images by using annotated words.

\subsection{Implementation Details}\label{sec:Implementation_Detail}
\textbf{Model Configurations}
The details of backbone are introduced in Sec.\ref{sec:backbone network}. For PVAM, the size of $W_e$, $W_o$ and $W_v$ is $512$, and the embedding dim of $f_o$ is $512$ in Eq.\ref{eq:attn}.
The embedding dim in GSRM is also set to $512$. 
The semantic reasoning block consists of 4 stacked transformer units, where the number of heads is $8$ and the number of hidden units is $512$.
For fair comparison, the same backbone as our SRN is adopted in the CTC, 1D-Attention and 2D-Attention based methods.
The number of both attention units and hidden units in 1D-Attention and 2D-Attention are set to $512$.

\textbf{Data Augmentation} 
The size of input images is $64\times 256$. We randomly resize the width of original image to 4 scales (e.g., 64, 128, 192, and 256), and then pad them to $64\times 256$.
Besides, some image processing operations~\cite{liao2019mask}, such as rotation, perspective distortion, motion blur and Gaussian noise, are randomly added to the training images.
The number of class is 37, including 0-9, a-z, and 'EOS'. And the max length of output sequence $N$ is set to $25$.

\textbf{Model Training} 
The proposed model is trained only on two synthetic datasets, namely Synth90K \cite{synth90k-1-Jaderberg14c,jaderberg2016-90K} and SynthText \cite{synthtext-Gupta16} without finetuning on other datasets. 
The ResNet50 pre-trained on ImageNet is employed as our initialized model and the batch size is $256$. 
Training is divided into two stages: warming-up and joint training. At first stage, we train the SRN without the GSRM for about 3 epochs. 
ADAM optimizer is adopted with the initial learning rate $1e^{-4}$. At joint training stage, we train the whole pipeline end-to-end with the same optimizer until convergence.
All experiments are implemented on a workstation with 8 NVIDIA P40 graphics cards.

\subsection{Ablation Study} \label{sec:Ablation_Study}

\subsubsection{Discussions about Backbone Network and PVAM}
Our SRN utilizes transformer units\cite{vaswani2017attention} in the backbone and adds the character reading order information in the PVAM to capture global spatial dependencies. 
As depicted in Tab.~\ref{tab:without_transformer},
our backbone with transformer units outperforms the one without it on all benchmarks by at least $3\%$ in accuracy, demonstrating that the importance of global visual context captured by transformer unit. 
As depicted in the in Tab.~\ref{tab:without_transformer}, the gain of using char reading order is obtained in the most public datasets, especially for CUTE.
The performance on some easy tasks is slightly improved, since the attention mechanism without this term is actually able to implicitly capture the reading order through data-driven training.

\begin{table}[htp]\footnotesize
  \begin{center}
      \caption{ 
         Ablation study of backbone and PVAM. ``Base'' means the backbone; ``TU'' means the transformer units; ``CRO'' means the character reading order information.}
      \label{tab:without_transformer}
      \begin{tabular}{|c|c|c|c|c|c|c|}
         \hline
         \makecell{Method}
          &IC13&IC15&IIIT5K&SVT&SVTP&CUTE\\
         \hline
         \makecell{Base}
          &90.0&72.4&87.0&83.6&73.8&80.5\\
         \makecell{Base+TU}
          &93.0&\textbf{77.5}&91.9&87.5&\textbf{79.8}&83.6\\
         \makecell{Base+TU+CRO} 
         &\textbf{93.2}&\textbf{77.5}&\textbf{92.3}&\textbf{88.1}&79.4&\textbf{84.7}\\
         \hline
      \end{tabular}
  \end{center}
  \vspace{-10mm}
\end{table}

\subsubsection{Discussions about GSRM} \label{sec:Discussions about GSRM}
To evaluate the effectiveness of GSRM in semantic reasoning, we compare the results yielded by the experiments with/without GSRM. Besides, the number of transformer units in GSRM is also explored. As shown in Tab.~\ref{tab:gsr}, the GSRM achieves successive gains of 1.5\%, 0.2\%, 0.8\%, 0.3\% in IIIT5K and 4.2\%, 0.9\%, 0.1\%, 0.0\% in IC15 with the number of Transformer Units set to 1, 2, 4, and 6. This suggests that the semantic information is important to text recognition and GSRM is able to take advantage of these information. Since the performance of 4-GSRM is similar to that of 6-GSRM, the 4-GSRM is adopted in the remaining experiments to preserve controllable computation. 
\begin{table}[htp]\footnotesize
  \begin{center}
      \caption{ 
         Ablation study of GSRM configuration. ``$n$-GSRM'' means the GSRM has $n$ transformer units.
      }
      \vspace{+1mm}
      \label{tab:gsr}
      \begin{tabular}{|l|c|c|c|c|c|c|}
         \hline
         &IC13&IC15&IIIT5K&SVT&SVTP&CUTE\\
         \hline
         \makecell{no GSRM} 
         &93.2&77.5 &92.3 &88.1 &79.4 &84.7\\
         \makecell{1-GSRM} 
         &94.7&81.7 &93.8 &88.5 &82.6 &\textbf{88.9}\\
         \makecell{2-GSRM} 
         &\textbf{95.6}&82.6 &94.0 &{91.0} &83.9 &87.8\\
         \makecell{4-GSRM} 
         &95.5&\textbf{82.7} &94.8 &\textbf{91.5} &\textbf{85.1} &87.8\\
         \makecell{6-GSRM} 
         &95.0 &\textbf{82.7} &\textbf{95.1} &90.6 &84&86.8\\
         \hline
      \end{tabular}
  \end{center}
  \vspace{-5mm}
\end{table}
\begin{table}[htp]\footnotesize
  \begin{center}
    \caption{Ablation study of semantic reasoning strategy. ``2D-ATT'' means 2D-Attention;  ``FSRM'' and ``BSRM'' mean forward and backward one-way semantic reasoning module respectively. 
         }
    \vspace{+1mm}
      \label{tab:global_superiority}
  \begin{tabular}{|l|c|c|c|c|c|c|}
         \hline
    &IC13&IC15&IIIT5K&SVT&SVTP&CUTE\\
  \hline
  CTC &91.7&74.6& 91.6& 84.5& 74.9& 81.2\\
  2D-ATT& 94.0&77.0&92.7&88.1&78.1&84.3\\
  FSRM& 94.7&81.1&\textbf{94.9}&89.6&81.7&87.1\\
  BSRM& 94.5&81.1&94.3 & 90.0& 82.5& 86.8\\
  GSRM& \textbf{95.5}&\textbf{82.7} &94.8 &\textbf{91.5} &\textbf{85.1} &\textbf{87.8}\\
  \hline
  \end{tabular}
  \end{center}
  \vspace{-8mm}
\end{table}

To demonstrate the benefits of global semantic reasoning strategy, we compare our approach with two variants: one only runs forward and the other runs backward to capture one-way semantic information. 
Moreover, the two typical text recognition methods, CTC and 2D-Attention based methods, are also included in the comparison to prove our superiority to both the existing semantic context-free methods and semantic context-aware methods. 
As illustrated in Tab.~\ref{tab:global_superiority}, all the semantic context-aware methods outperform the semantic context-free methods (CTC based methods), which highlights the importance of semantic information. Furthermore, the GSRM with global semantic reasoning outperforms those with the one-way semantic reasoning by about 1\% in accuracy on most of the benchmarks, verifying the effectiveness of the multi-way semantic reasoning.
\subsubsection{Discussions about Feature Fusion Strategy}
In this paper, we introduce a novel feature fusion strategy, namely gated unit, which is described in Sec.~\ref{sec:Visual-Semantic Fusion Decoder}. In this section, we conduct experiments to compare our VSFD with three different feature fusion methods, including add, concatenate and dot. Tab.~\ref{tab:fusion} indicates that the other three fusion operations will lead to degradation of performance on benchmarks to a certain extent. Thus, the VSFD is utilized in our approach as default.

\begin{table}[htp]\footnotesize
  \begin{center}
    \caption{Ablation study of feature fusion strategy.}  
  \label{tab:fusion}
  \begin{tabular}{|l|c|c|c|c|c|c|}
         \hline
    &IC13&IC15&IIIT5K&SVT&SVTP&CUTE\\
  \hline
  Add & 95.2&81.7& 93.8& 90.9& 84.3& 87.8\\
  Concat & 95.0 &82.0& 93.8 & \textbf{91.5}& 82.9& \textbf{88.1}\\
  Dot & 94.8 & 81.0 & 92.0& 89.7& 84.5& \textbf{88.1}\\
  \makecell[c]{Gated unit} & \textbf{95.5}& \textbf{82.7}&\textbf{94.8}& \textbf{91.5}&\textbf{85.1} & 87.8\\
  \hline
  \end{tabular}
  \end{center}
  \vspace{-9mm}
\end{table}
\subsubsection{Analysis of Right/Failure Cases}
\begin{figure}[htp]
\begin{center}
      \includegraphics[width=0.8\columnwidth]{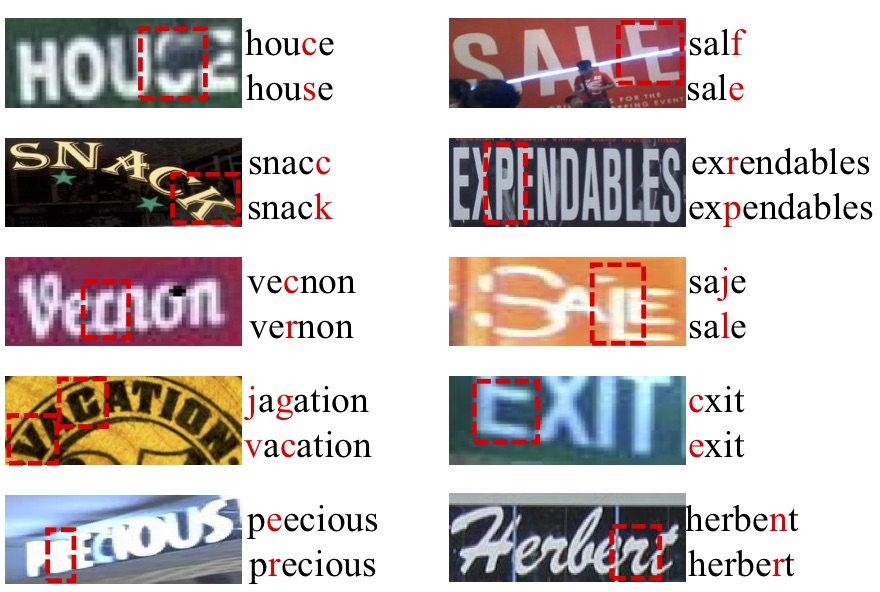}
  \caption{Right cases of SRN with/without GSRM. The predictions are placed along the right side of images.
  The top string is the prediction of SRN without the GSRM; The bottom string is the prediction of SRN.
  }
  \label{fig:right-case} 
\end{center}
\vspace{-5mm}
\end{figure}

To illustrate how semantic information helps SRN to improve the performance, we collect some individual cases from the benchmarks to compare the predictions of SRN with/without GSRM. As shown in Fig.~\ref{fig:right-case}, for example, because the character “r” is visually similar to character “c” in the image with word “Vernon”, the prediction without GSRM wrongly gives the character “c”, while the prediction with GSRM correctly infers the character “r” with the help of global semantic context. 
The character “e” in “sale”, the character “r” in “precious”, and the character “n” in “herbert” are handled by the same working pattern.
\begin{figure}[htp]
  \begin{center}
      \includegraphics[width=0.7\columnwidth]{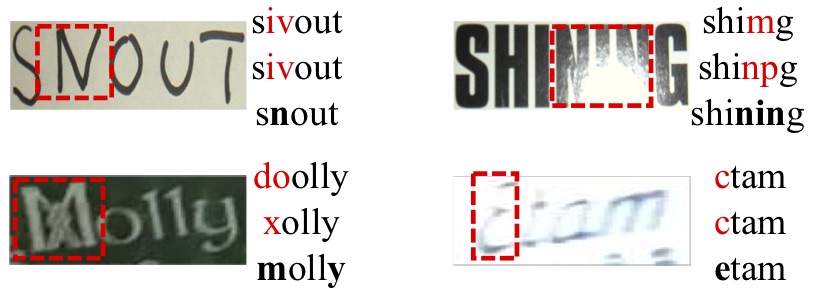}
   
  \caption{ Failure cases of SRN.
The predictions and ground-truth labels are placed along the right side of images. 
The top string is the prediction of SRN without the GSRM;
The middle string is the prediction of SRN;
The bottom string is the ground-truth.
  }
  \label{fig:bad-case} 
\end{center}
\vspace{-8mm}
\end{figure}

The failure cases of SRN is shown in Fig.~\ref{fig:bad-case}, which can be divided into two categories: 1) SRN cannot insert or remove characters, but only modifies wrong characters. If the visual features are aligned wrongly or some characters are missed, SRN cannot infer the correct characters, such as “snout” and “shining”. 
2) The SRN may not work when both visual and semantic context fail to recognize the image, since the gain of SRN is mainly due to the complementation in visual features and semantic features.
When the image suffers from special fonts or low quality and the words in it are rarely appeared in training data, it is difficult for the SRN to get the visual context and semantic dependencies, as the second line in Fig.~\ref{fig:bad-case} shows.

\subsection{Comparisons with State-of-the-Arts}\label{sec:Comparsion_on_Benchmark}
The comparison of our method with previous outstanding methods is shown in Tab.~\ref{tab:benchmark}. We only compare the results without any lexicon, because the lexicon is always unknown before recognition in practical use. The context-aware methods perform better than context-free methods in general, and the proposed SRN achieves superior performance across the six public datasets compared with state-of-the-art approaches with the help of GSRM, which proves that this particular designed module can make better use of semantic information than existing techniques. For regular datasets, we get a 0.2\%, 0.4\%, 0.9\% improvement on IC13, IIIT5K and SVT respectively. The gain of SVT is quite larger than the other two sets, and we claim that semantic information will play a more significant role, especially for recognition of the low-quality images.

Although our method does not take special measures into consideration, such as rectification module, to handle the irregular words, like ASTER~\cite{shi2018aster} and ESIR~\cite{zhan2019esir}, it is worth noting that the SRN achieves comparable or even better performance on those distorted datasets. As is shown in Tab.~\ref{tab:benchmark}, there are increase of 4.0\% and 2.8\% on IC15 and SVTP respectively and comparable results on CUTE, comparing with State-of-the-Arts methods with rectification modules. Similar to the explanation of the gain on SVT, we believe this is mainly due to the fact that global semantic information will be an important supplementation to the visual information in text recognition, and it will show more effectiveness when facing difficult cases. 
\begin{table*}[htp]\footnotesize
  \begin{center}
      \caption{Comparisons of scene text recognition performance  with previous methods on several benchmarks. 
      All results are under NONE lexicon.
      ``90K'' and ``ST'' mean Synth90K and SynthText; ``word'' and ``char''
      means the word-level or character-level annotations are adopted; ``self'' means 
      self-designed convolution network or self-made synthetic datasets are used. 
      SRN w/o GSRM means that SRN cuts down GSRM, and thus loses semantic information.}
      \label{tab:benchmark}
  \begin{tabular}{|c|l|l|l|c|c|c|c|c|c|}
  \hline
  &Method &ConvNet,Data &Annos&IC13&IC15&IIIT5K&SVT&SVTP&CUTE\\
  \hline
   \multirow{5}{*}{\makecell[c]{ Semantic \\context\\-free}}
  &Jaderberg~\etal~\cite{jaderberg2016-90K}&VGG,90K&word&90.8 &-&-&80.7&-&-\\
  &Jaderberg~\etal~\cite{jaderberg2014deep-structuredoutput}&VGG,90K&word&81.8 &-&-&71.7&-&-\\
  &Shi~\etal~\cite{crnn-shi2016end}~(CTC)&VGG,90K&word&89.6&-&81.2&82.7&-&-\\
  &Lyu~\etal~\cite{lyu20192d}~(Parallel)&ResNet,90K+ST&word& 92.7 &76.3 &94.0 &90.1 & 82.3& 86.8\\
  &Xie~\etal~\cite{xie2019aggregation}~(ACE)&VGG,90K&word &89.7& 68.9& 82.3 & 82.6 & 70.1& 82.6\\
  &Liao~\etal~\cite{liao2019scene-fcn}~(FCN)&ResNet,ST&word,char&91.5&-&91.9&86.4&-&-\\
  \hline
   \multirow{10}{*}{\makecell[c]{ Semantic \\context\\-aware}}
  &Lee~\etal~\cite{lee2016recursive}&VGG,90K&word&90.0&-&78.4&80.7&-&-\\
  &Cheng~\etal~\cite{cheng2017focusing}~(FAN)&ResNet,90k+ST&word&93.3&70.6&87.4&85.9&-&-\\
  &Cheng~\etal~\cite{cheng2018aon}~(AON)&self,90k+ST&word&-&68.2&87.0&82.8&73.0&76.8\\
  &Bai~\etal~\cite{bai2018edit}&ResNet,90K+ST&word&94.4&73.9&88.3&87.5&-&-\\
  &Yang~\etal~\cite{yang2017learning-2D}&VGG,90K+self&word,char&- & - & -& -& 75.8 & 69.3\\
  &Shi~\etal~\cite{shi2018aster}~(ASTER)&ResNet,90K+ST&word& 91.8 &76.1 &93.4 &89.5 &78.5 &79.5\\
  &Zhan~\etal~\cite{zhan2019esir}~(ESIR)&ResNet,90K+ST&word&  91.3&76.9 &93.3 &90.2 &79.6 &83.3 \\
  &Yang~\etal\cite{yang2019symmetry}~(ScRN)&ResNet,90K+ST&word,char& 93.9& 78.7 & 94.4 & 88.9   & 80.8 & 87.5\\
  &Li~\etal~\cite{SAR-li2019show}~(SAR)&ResNet,90K+ST&word&91.0&69.2& 91.5 &84.5 &76.4 &83.3\\
  &Liao~\etal~\cite{liao2019mask}~(SAM) &ResNet,90K+ST&word& 95.3 &77.3 &93.9 &90.6 &82.2 &\textbf{87.8}\\
  \hline
  \multirow{2}{*}{Ours}
  &SRN w/o GSRM                 &ResNet,90K+ST&word& 93.2 &77.5 &92.3 &88.1 &79.4 &84.7\\
  &SRN            &ResNet,90K+ST&word& \textbf{95.5} &\textbf{82.7} &\textbf{94.8} &\textbf{91.5} &\textbf{85.1} &\textbf{87.8}\\
  \hline
  \end{tabular}
  \end{center}
  \vspace{-6mm}
\end{table*}

\subsection{Results on non-Latin Long Text}
To evaluate the performance on long text, we set up two additional experiments: Attention and CTC based methods with the same configuration. We generate a synthetic long text dataset with the engine in~\cite{synthtext-Gupta16}, which includes 3 million images. Besides, we also use the training set of RCTW~\cite{shi2017icdar2017} and LSVT~\cite{sun2019chinese} as training data. Following the configuration described in Sec.~\ref{sec:Implementation_Detail}, we just change the max decoding length $N$ to 50 and the number of classes to 10784. We evaluated our model on ICDAR2015 Text Reading in the Wild Competition dataset (TRW15)~\cite{zhou2015icdar} by character-level accuracy. TRW15 dataset contains 484 test images. We crop 2997 horizontal text line images as the first test set (TRW-T) and select the images whose length is more than 10 as the second test set (TRW-L). 

The results are shown in Tab.~\ref{tab:longtext}. Compared with CTC and attention based methods, the proposed approach without GSRM achieved 6.8\% and 8.4\% boost in TRW-T. Because our method could model 2D spatial information and conquer the error accumulation when the wrong decoding is raised at a certain time step. Compared with SCCM~\cite{yin2017scene}, our SRN achieves 4.9\% improvement over the SRN without GSRM, while LM model in SCCM obtains 4.7\% improvement. This shows that GSRM could integrate semantic features with visual features very well, which is important to recognition long text. Compared with the accuracy of TRW-T and TRW-L in the table, the improvement of GSRM increases from 4.9\% to 6.8\%. We can also find that 2D-Attention has a much lower recognition rate in the TRW-L, which is approximately caused by the accumulation of errors 

\begin{figure}
    \centering
    \includegraphics[width=0.8\columnwidth]{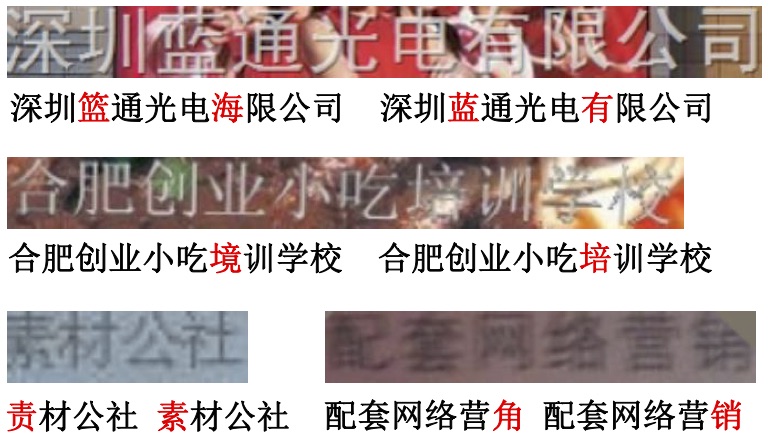}
    \caption{Right cases for non-Latin long text. There are two predictions under the image. 
    The left string is the prediction of SRN without the GSRM; 
    The right string is the prediction of SRN.}
    \label{fig:right_case_longtext}
    \vspace{-1mm}
\end{figure}

As shown in Fig.~\ref{fig:right_case_longtext}, there are several cases selected from the test set. It's obvious that semantic information can better distinguish two characters, when they are easily confused. For example, "责" is visually similar to "素", while "素材" is a common Chinese phrase, so the SRN with GSRM correctly infer the character.

\begin{table}[htp]\footnotesize
  \begin{center}
  \caption{Recognition accuracies (\textbf{character-level}) on non-Latin long text dataset}
  \label{tab:longtext}
  \begin{tabular}{|l|c|c|}
  \hline
  Method &TRW-T(\%) & TRW-L(\%)\\
  \hline
  CASIA-NLPR\cite{zhou2015icdar}& 72.1 &-\\
  SCCM w/o LM\cite{yin2017scene}& 76.5 &-\\
  SCCM\cite{yin2017scene}& 81.2 &-\\
   
  2D-Attention& 72.2 &59.8\\
  CTC& 73.8 & 70.9\\
  \hline
  SRN w/o GSRM& 80.6 & 77.5\\
  SRN& \textbf{85.5} & 84.3\\
  \hline
  \end{tabular}
  \end{center}
\vspace{-6mm}
\end{table}

\subsection{Inference Speed}
\begin{table}[htp]\footnotesize
  \begin{center}
  \caption{Comparion of speed during inference} 
      \label{tab:speed}
  \begin{tabular}{|l|c|c|}
  \hline
    Method &\makecell{IC15}& TRW-L\\
  \hline
    CTC& 128.6ms &  131.8ms \\
    1D-Attention& 323.3ms & 431.1ms \\
    2D-Attention& 338.8ms & 486.9ms \\
    \hline
   
    SRN w/o GSRM& 131.5ms & 137.3ms \\
    SRN & 191.6ms  & 216.8ms \\
   
  \hline
  \end{tabular}
  \end{center}
  \vspace{-7mm}
\end{table}
To explore the efficiency of our proposed approach, we evaluate the speed of our method with/without GSRM and compare it with CTC, 1D-Attention and 2D-Attention based recognizers in both the short and long text datasets. The test set is IC15 and TRW-L, of which the average length is 5 and 15 respectively. 
For a fair comparison, we test all methods with the same backbone network on the same hardware (NVIDIA Tesla K40m). Each method runs 3 times on the test set, and the average time consumed by a single image is listed in Tab.~\ref{tab:speed}. 

Benefiting from the parallel framework in SRN, our model with GSRM is 1.7 times and 1.8 times faster than the 1D and 2D-Attention based method in the IC15, and the acceleration will be enlarged to 2.0 times and 2.2 times in the TRW-L.
Meanwhile, the computational efficiency of our approach without GSRM is similar to that of CTC-based method, due to its parallelism and simplicity. 
\section{Conclusion}
In this paper, we claim that semantic information is of great importance for a robust and accurate scene text recognizer.
Given the characters (Latin or non-Latin) of a text line, we use the GSRM to model its semantic context, which includes both first-order relations among characters and higher-order relations.
Integrating with GSRM, we propose a novel end-to-end trainable framework named semantic reasoning network (SRN) for recognizing text in the wild, which also contains backbone network, parallel visual attention module and fusion decoder module. 
SRN achieves SOTA results in almost 7 public benchmarks including regular text, irregular text and non-Latin long text, and extensive experiments are conducted to show the significant superiority over the existing methods. 
Additionally, since all modules of SRN are time independent, SRN can run in parallel and is more practical than other semantic modeling methods. 
In the future, we are interested in improving the efficiency of GSRM, and making it adaptive to CTC-based methods to boost its value in practical applications.

{\small
\bibliographystyle{ieee_fullname}
\bibliography{egbib}
}
\end{CJK}
\end{document}